\documentclass{article}
\usepackage{multirow}
\usepackage{longtable}
\usepackage{arxiv}
\usepackage{pdflscape}
\usepackage{subfig}
\usepackage{tabularx} 
\usepackage{booktabs}
\usepackage[per-mode=symbol]{siunitx}
\usepackage[x11names]{graphicx}

\usepackage[utf8]{inputenc} 
\usepackage[T1]{fontenc}    
\usepackage[colorlinks=true, linkcolor=blue, citecolor=blue, urlcolor=blue, filecolor=blue, pdfborder={0 0 0}]{hyperref}
\usepackage{url}            
\usepackage{booktabs}       
\usepackage{amsfonts}       
\usepackage{nicefrac}       
\usepackage{microtype}      
\usepackage{lipsum}		
\usepackage{graphicx}
\usepackage{doi}
\usepackage{amsmath}
\usepackage{geometry}
\usepackage{array}
\usepackage{cleveref}
\usepackage{float}
\usepackage{lscape}
\usepackage{makecell}
\usepackage{adjustbox}
\usepackage{authblk}  
\usepackage{amssymb } 

\usepackage[x11names]{xcolor}
 
\extrafloats{500}
 \title{Social LSTM with Dynamic Occupancy Modeling for Realistic Pedestrian Trajectory Prediction
 }

 \author{Ahmed Alia\href{https://orcid.org/0000-0002-3049-4924}{\includegraphics[scale=0.06]{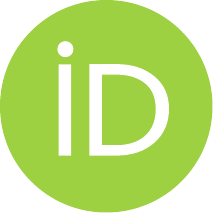}}$^{1,3}$, 
Mohcine Chraibi
\href{https://orcid.org/0000-0002-0999-6807}{\includegraphics[scale=0.06]{orcid.pdf}}$^{1}$, 
Armin Seyfried\href{https://orcid.org/0000-0001-8888-0978} 
{\includegraphics[scale=0.06]{orcid.pdf}}$^{1,2}$ } 

\affil{
  $^{1}$ Institute for Advanced Simulation, Forschungszentrum Jülich, 52425 Jülich, Germany 
}
\affil{
  $^{2}$ Faculty of Architecture and Civil Engineering, University of
 Wuppertal, 42285 Wuppertal, Germany
}
\affil{
  $^{3}$ Faculty of Information Technology and Artificial Intelligence, An-Najah National University, P4110257 Nablus, Palestine 
}

\renewcommand{\shorttitle}{\textit{} }

\hypersetup{
pdftitle={A template for the arxiv style},
pdfsubject={q-bio.NC, q-bio.QM},
pdfauthor={David S.~Hippocampus, Elias D.~Striatum},
pdfkeywords={Deep Learning, Computer Vision, Object Detection},
}

\definecolor{DarkGreen}{RGB}{0,100,0}

\begin{document}
\maketitle

\begin{abstract}

 In dynamic and crowded environments, realistic pedestrian trajectory prediction remains a challenging task due to the complex nature of human motion and the mutual influences among individuals. Deep learning models have recently achieved promising results by implicitly learning such patterns from 2D trajectory data. However, most approaches treat pedestrians as point entities, ignoring the physical space that each person occupies.
 To address these limitations, this paper proposes a novel deep learning model that enhances the Social LSTM with a new Dynamic Occupied Space loss function. This loss function guides Social LSTM in learning to avoid realistic collisions without increasing displacement error across different crowd densities, ranging from low to high, in both homogeneous and heterogeneous density settings. Such a function achieves this by combining the average displacement error with a  new collision penalty that is sensitive to scene density and individual spatial occupancy. For efficient training and evaluation, five datasets were generated from real pedestrian trajectories recorded during the Festival of Lights in Lyon 2022. Four datasets represent homogeneous crowd conditions—low, medium, high, and very high density—while the fifth corresponds to a heterogeneous density distribution. 
 The experimental findings indicate that the proposed model not only lowers collision rates, but also enhances displacement prediction accuracy in each dataset. Specifically, the model achieves up to a \SI{31}{\percent} reduction in the collision rate and reduces the average displacement error and the final displacement error by  \SI{5}{\percent} and  \SI{6}{\percent}, respectively, on average across all datasets compared to the baseline. Moreover, the proposed model consistently outperforms several state-of-the-art deep learning models across most test sets.
 
\end{abstract}

\section{Introduction}

The accurate and realistic prediction of pedestrian trajectories in crowded environments is crucial for a wide range of applications, including robotic navigation~\cite{sang2023rdgcn}, urban planning~\cite{tamaru2024enhancing}, and crowd management systems~\cite{zong2024pedestrian}. Despite significant progress, this task remains challenging in dense crowds due to the complexity of the dynamic nature of pedestrian behavior~\cite{wang2025pedestrian, xie2024pedestrian, mohamed2020social}. 
In such environments, individuals continually adjust their paths in response to their surroundings, especially in reaction to the movements of others.
Early research addressed these challenges using models based on physics or models inspired by robotics, such as the social force model~\cite{helbing1995social, gil2023human}, cellular automata model~\cite{burstedde2001simulation}, and the optimal reciprocal collision avoidance~\cite{van2011reciprocal}.
These methods, grounded in well-defined rules or differential equations, established a strong foundation for understanding pedestrian movement and continue to provide valuable insights~\cite{korbmacher2022review, chatagnon2025exploring}.
With advances in deep learning and its success in various human crowd analysis tasks~\cite{abubaker2025rpee, alia2024novel}, deep learning algorithms have gained increasing attention for pedestrian trajectory prediction. These methods have demonstrated reliable performance in forecasting several seconds into the future (e.g., \SI{4}{\second})~\cite{sighencea2021review, yang2024ia, jiang2025survey}.
Among the most widely used deep learning methods are Long Short-Term Memory networks (LSTM)~\cite{hochreiter1997long} and Generative Adversarial Networks (GAN)~\cite{goodfellow2014generative}.  Although these models effectively learn and predict movement sequences, they often treat each pedestrian's path independently, overlooking the influence of nearby individuals. This limits their ability to produce accurate and realistic forecasts in crowded environments.

To address this limitation, advanced models such as Social-LSTM~\cite{alahi2016social} and Social-GAN~\cite{gupta2018social} incorporate mechanisms that account for the influence of nearby movements using 2D trajectories of multiple pedestrians. Social-LSTM, for example, assigns an individual LSTM to each pedestrian and introduces a social pooling layer that aggregates the hidden states of neighbors within a local spatial grid. This design allows the model to adjust the predictions based on the motion of the surrounding individuals. 
To avoid confusion, the term ``social'' in these models does not denote genuine social interaction in a semantic sense, since 2D trajectory data alone cannot capture behaviors that depend on non-kinematic cues such as intentions, gaze, gestures, or verbal communication~\cite{hasan2019forecasting, sadeghian2017tracking}. 

Although these advances improve the modeling of pedestrian motion, they often fall short in ensuring that the predicted trajectories remain physically realistic, particularly in dense crowds. A key limitation is that pedestrians are represented as single points, which neglects their physical dimensions and spatial occupancy in the environment, often leading to unrealistic collisions~\cite{chatagnon2025exploring, awan2023trajectory}. 
Recently, a metric has been proposed to evaluate a model's ability to generate collision-free paths between the bodies of pedestrians, representing each individual as a circular disk with a fixed radius of \SI{0.2}{\meter} to approximate occupied space~\cite{\cite{kothari2021human}}.   
Building on this motivation, a new loss function for Social LSTM was introduced that not only minimizes displacement errors, but also explicitly learns collision avoidance~\cite{korbmacher2024toward}. 
While effective in low- and medium-density environments, this approach underperforms in highly dense and heterogeneous crowds.

To overcome these limitations, this paper proposes a novel deep learning model based on Social-LSTM, enhanced by a new Dynamic Occupied Space (DOS) loss function. The proposed model leverages this loss to explicitly learn realistic collision avoidance while also improving displacement accuracy across various crowd densities, both homogeneous and heterogeneous.

The contribution of this paper can be summarized as follows.
\begin{enumerate}
    \item This paper presents a novel deep learning model that combines the Social LSTM architecture with a new loss function to predict pedestrian trajectories more accurately and realistically in diverse and dynamic crowd environments.
    \item It introduces a collision-aware loss function based on a circular disk with a dynamic radius that adapts to scene density, enabling Social-LSTM to better capture physical realism while improving displacement accuracy.
    \item The study uses five datasets that represent low, medium, high, and very high crowd densities, along with a heterogeneous dataset that includes a mix of all four. The data source for these datasets is real pedestrian trajectories recorded during Lyon's Festival of Lights in 2022.
\end{enumerate}

The remainder of this paper is organized as follows. Section~\ref{sec:realtedwork} reviews existing studies on the prediction of pedestrian trajectory. 
Section~\ref{sec:methodology} outlines the methodology behind the proposed deep learning framework to predict realistic pedestrian trajectory in crowded environments.
Section~\ref{sec:experimentalandresults} presents the experimental setup, results, and comparative analysis.
Finally, Section~\ref{sec:conclusion} summarizes the key findings, discusses limitations, and outlines potential directions for future research.

\section{Related Work}
\label{sec:realtedwork}

Efficient prediction of pedestrian trajectory in crowded environments remains a challenging task, particularly due to the need to model nearby movements and ensure physically plausible, collision-free paths. Recent deep learning methods have significantly advanced the field by learning from large-scale trajectory datasets~\cite{wang2024pedestrian, yang2025pedestrian, chen2025trajectory}.

\subsection{Neighbor-based Models}

In this category, models leverage the movements of nearby pedestrians to account for the influence of surrounding trajectories.
Early models such as Social LSTM~\cite{kothari2021human} introduced social pooling to implicitly model interactions between each individual and nearby pedestrians. Based on this, SR-LSTM~\cite{zhang2019sr} and STGAT~\cite{huang2019stgat} incorporated attention mechanisms and spatio-temporal graph structures to capture richer interaction patterns. To broaden the spatial context, Xue et al.~\cite{xue2019location} proposed a joint Location–Velocity Attention LSTM model for pedestrian trajectory prediction, where an attention mechanism learns to combine location and velocity cues to improve forecasting accuracy and generalization across scenes.

More recent architectures, such as Social-BiGAT~\cite{kosaraju2019social} and AgentFormer~\cite{yuan2021agentformer}, employed attention-based frameworks—Social-BiGAT used a graph attention network for social interactions, while AgentFormer leveraged a transformer to jointly capture temporal and social dependencies. Advancements such as SocialFormer~\cite{li2024iaha} and Knowledge-Aware Graph Transformer~\cite{zhu2024propagation} extended attention-based frameworks by incorporating semantic interaction features and adaptive mechanisms to better capture complex relational dependencies.

\subsection{Generative-based Models}
Generative models have emerged to address the multimodal and uncertain nature of human behavior by generating multiple plausible future trajectories. Social GAN~\cite{gupta2018social} was among the first to apply a generative adversarial framework, combining LSTM-based encoders and decoders with a discriminator to produce diverse and socially compliant paths.  Trajectron++~\cite{salzmann2020trajectron} proposed a graph-structured recurrent model to forecast multi-agent trajectories by incorporating agent dynamics and environmental context.

More recent approaches, such as GSGFormer~\cite{luo2023gsgformer}, integrate graph attention with CVAE modules for semantically consistent generation, while TUTR~\cite{song2023tutr} simplifies the process by using a transformer-based framework to directly predict trajectory modes and their probabilities.

\subsection{Physical Realism and Collision Avoidance}

Despite these advances, most of the existing deep learning–based trajectory prediction approaches still model pedestrians as abstract points, ignoring their physical space requirements. This limits their ability to ensure realistic and collision-free predictions in dense environments~\cite{chatagnon2025exploring, song2019experiment}.

To address this limitation, recent research has incorporated representations of personal body space and penalized overlapping trajectories. For example, Time-to-Collision (TTC)-Social LSTM~\cite{korbmacher2024toward} uses a TTC-based loss function and represents each pedestrian as a circular disk with a fixed radius of \SI{0.2}{\meter}. This design helps the model learn realistic collision avoidance behavior while maintaining displacement accuracy, thereby improving realism and predictive accuracy in crowded scenes. However, TTC-Social LSTM still faces challenges in dense and heterogeneous crowds, where avoiding collisions can reduce the accuracy of the prediction. One contributing factor is the use of a single fixed body-space size for all pedestrians, regardless of scene density.

As a result, this article proposes a novel deep learning model that integrates the Social-LSTM architecture with a dynamic loss function of the occupied space, which adapts the space size of each person according to the density of the scene. This approach is designed to reduce collision occurrences and displacement errors in various conditions, including homogeneous low-, medium-, and high-density scenes, as well as heterogeneous density distributions.
The following section presents a detailed overview of the proposed model.

\section{Methodology}
\label{sec:methodology}

This section begins with the pedestrian trajectory prediction problem. Then it introduces the proposed model that predicts realistic human trajectories. Finally, it presents the evaluation metrics used to assess the model.

\subsection{Problem Statement }
\label{sec:problemstatement}

\begin{figure} 
\centering
\includegraphics[width=1\linewidth]{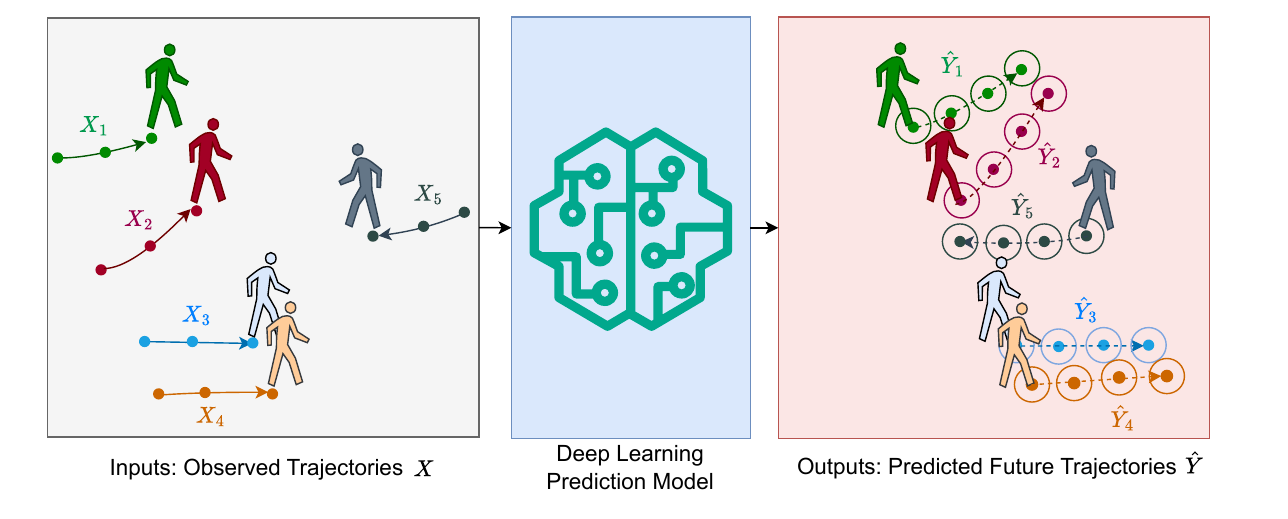}
    \caption{Illustration of the pedestrian trajectory prediction problem, where the goal is to minimize displacement error and avoid collisions between the predicted future paths. The circles with a radius of  \SI{0.2}{\meter} represent the body of a person and its space requirement.}
    \label{fig:problem}
\end{figure}

As shown in~\cref{fig:problem}, the proposed model processes the observed trajectories $X$ of all individuals $n$ in a scene, formulated as
$X = \{X_1, X_2, \dots, X_n\}$,
and predicts their future trajectories $\hat{Y}$, given by
$\hat{Y} = \{\hat{Y}_1, \hat{Y}_2, \dots, \hat{Y}_n\}$. Each observed trajectory $X_i$ of the pedestrian $i$ over time steps $t$ from $t_1$ to $t_{obs}$ is expressed as:

\begin{equation}
    X_i = \{(x_i^t, y_i^t) \mid t=t_1,t_2,\dots,t_{obs}\},
\end{equation}

where \( t_1 \) refers to the first observed time step, \( t_{\text{obs}} \) denotes the final observed time step, and \( (x_i^t, y_i^t) \) represents the spatial coordinates of the pedestrian \( i \) in the time step \( t \).
Similarly, the future trajectory $\hat{Y}_i$ predicted of individual $i$ over time steps $t$ from $t_{obs+1}$ to $t_{pred}$ is defined as:

\begin{equation}
    \hat{Y}_i = \{(\hat{x}_i^t, \hat{y}_i^t) \mid t=t_{obs+1}, t_{obs+2}, \dots, t_{pred}\},
\end{equation}

where $t_{pred}$ represents the final time step for the future trajectory predicted by the person $i$.
Furthermore, the ground truth trajectories corresponding to $\hat{Y}$ can be defined as
$Y = \{Y_1, Y_2, \dots, Y_n\}$, where $Y_i$ is given by:

\begin{equation}
    Y_i = \{(x_i^t, y_i^t) \mid t=t_{obs+1}, t_{obs+2}, \dots, t_{pred}\}.
\end{equation}

In addition, the observed trajectory \( X_i \) and the future trajectory \( Y_i \) together form the composed trajectory \( a_i(t_1) \) of the person \( i \) within a scene, starting from the time step \( t_1 \) and ending at \( t_{\text{pred}} \). The trajectory \( a_i(t_1) \) is defined as:

\begin{equation}
    a_i(t_1) = \{(x_i^t, y_i^t) \mid t = t_1, t_2, \dots, t_{\text{pred}}\}.
\end{equation}

To determine whether a collision occurs between two pedestrians $i$ and $j$ at the time step $t$, each pedestrian is modeled as a circle with a fixed radius of $R = 0.2$ m~\cite{yan2025generating, kothari2021human, korbmacher2024toward}. A collision is detected if their respective circles intersect. Mathematically, this condition is evaluated using the Euclidean distance between the centers of the two pedestrians, which is expressed as:

\begin{equation}
\text{collision}(i, j, t) =
\begin{cases}
1 & \text{if } \sqrt{(x_i^t - x_j^t)^2 + (y_i^t - y_j^t)^2} < 2R, \\
0 & \text{otherwise},
\end{cases}
\label{eq:collision}
\end{equation}
the collision indicator function returns 1 if a collision is detected and 0 otherwise.

\subsection{Dynamic Occupied Space-based Social LSTM Model}
\subsubsection{Social LSTM}
\begin{figure} 
\centering
\includegraphics[width=1\linewidth]{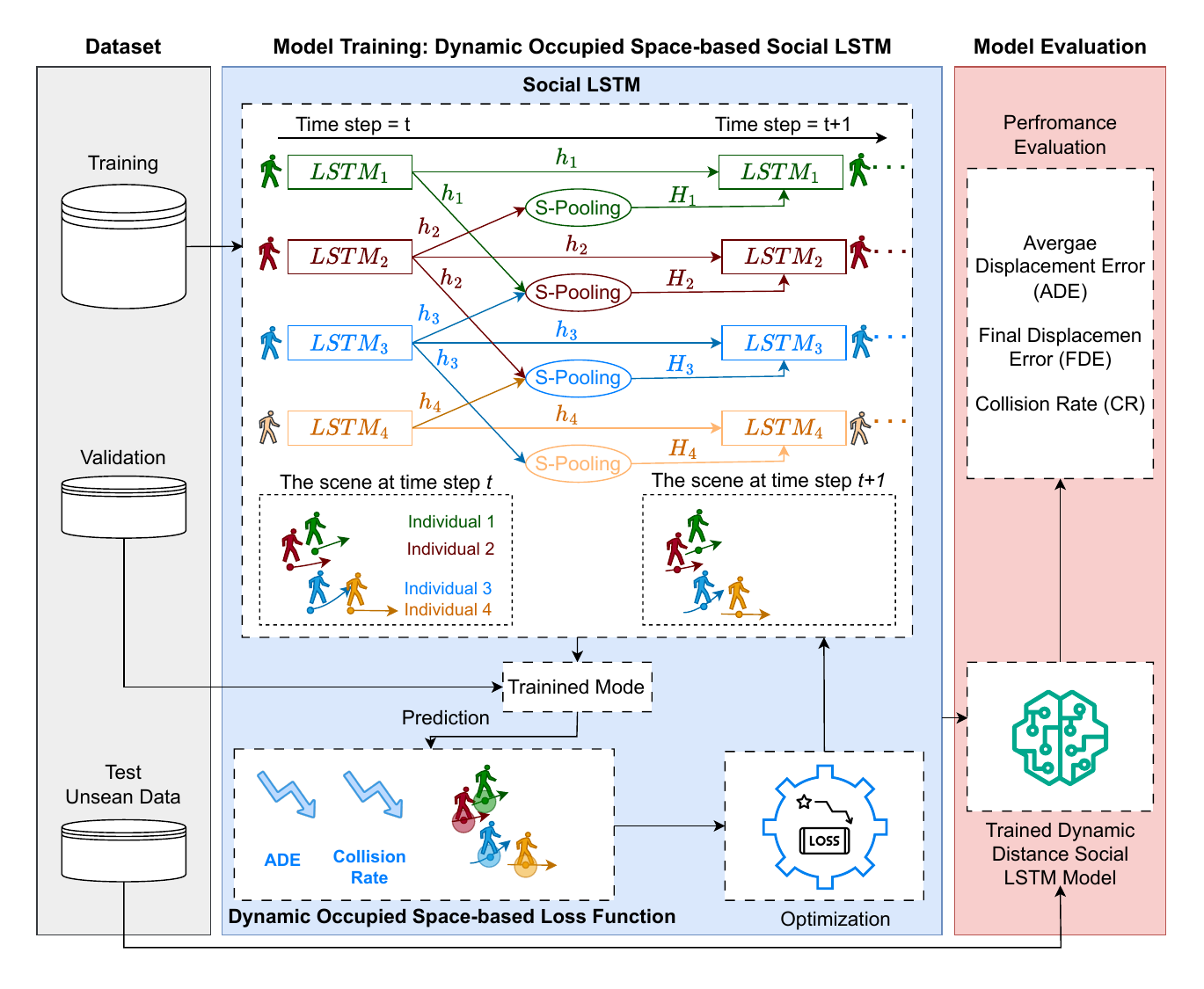}
    \caption{Overview of our model, including its training and evaluation process. In Social LSTM, S-Pooling refers to social pooling layers that share hidden states ($h$) among nearby LSTMs, while circles represent the occupied space of individuals.}
    \label{fig:methodology}
\end{figure}

Social LSTM is a recurrent neural network architecture designed to predict the future trajectories of individuals in crowded spaces. As illustrated in~\cref{fig:methodology}, the model extends the traditional LSTM by introducing a social pooling layer, which allows neighboring LSTMs to share their hidden states that encode time-varying motion patterns. In this architecture, each pedestrian is modeled by a separate LSTM that processes a sequence of their observed positions over time. To explicitly capture interactions among pedestrians, the social pooling layer aggregates the hidden states of pedestrians located within a predefined spatial neighborhood at each time step.
For example, in~\cref{fig:methodology}, $LSTM_1$ and $LSTM_3$ share their hidden states ($h_1$ and $h_3$) with $LSTM_2$ via its pooling layer, since individuals 1 and 3 fall within the predefined neighborhood of person 2. The aggregated features ($H_2$) from the pooling layer, combined with the hidden internal state of $LSTM_2$ ($h_2$), are then used to predict the future trajectory of individual 2. In contrast, $LSTM_2$ does not share its hidden state ($h_2$) with$LSTM_4$, as individual 4 lies outside the predefined spatial threshold. For more details about Social LSTM, the reader is referred to~\cite{alahi2016social}.

Despite its ability to reduce displacement errors in crowded scenes, Social LSTM occasionally fails to prevent pedestrian collisions on the predicted trajectories~\cite{kothari2021human}. This limitation arises because the model is trained to minimize displacement error without explicitly considering collision avoidance~\cite{kosaraju2019social}. As a result, the notion of collision avoidance is expected to be learned implicitly. However, this implicit modeling can lead to unrealistic trajectory predictions, particularly in dense crowd scenarios.
Therefore, the original Social LSTM model should be enhanced to train not only to minimize displacement error but also to explicitly avoid collisions.

\subsubsection{Dynamic Occupied Space-based Loss Function}
Here, we propose Dynamic Occupied Space ($\mathcal{L}_{DOS}$), a new and efficient loss function for Social LSTM that encourages collision avoidance and reduces displacement errors. This leads to more realistic crowd movement predictions across scenarios with different densities.

The proposed loss function combines the Average Displacement Error (ADE) with a new Collision Penalty ($\mathcal{CP}$) metric, and is defined as:

\begin{figure}
\centering
\includegraphics[width=1\linewidth]{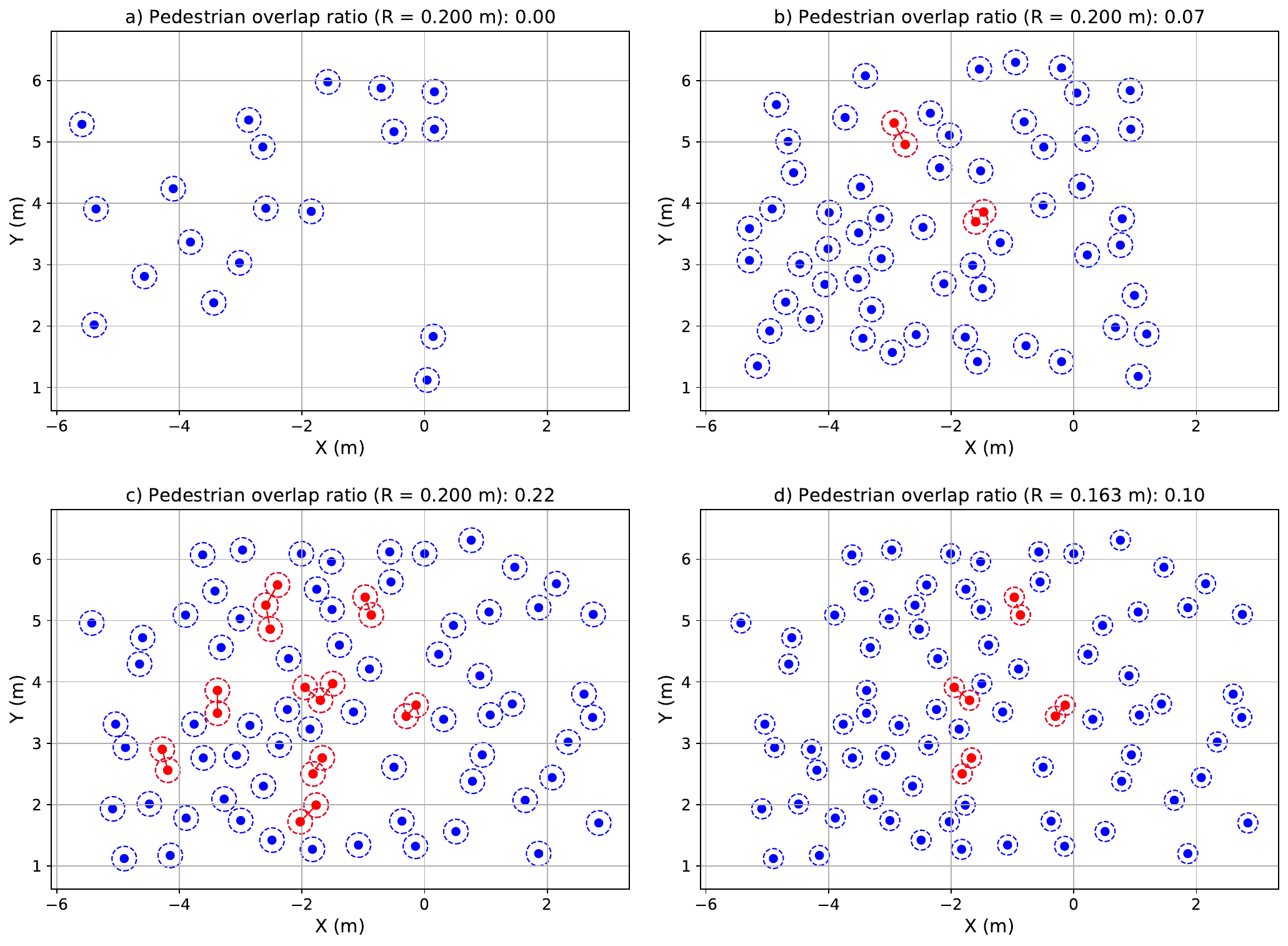}
    \caption{Visualized examples of pedestrians within a scene at a specific frame, illustrating varying overlap ratios. Each circle with radius $R$ represents the physical space occupied by an individual. Red circles indicate pedestrians involved in collisions, with red lines showing the distance between the centers of the colliding individuals. Blue circles represent pedestrians without any overlaps. }
    \label{fig:radiusexamples}
\end{figure}

\begin{equation}
\mathcal{L}_{\text{DOS}}=ADE +\lambda \times \mathcal{CP}, 
\label{eq:DOC-SLSTM}
\end{equation}

where $\lambda$ is a predefined weight that controls the influence of the $\mathcal{CP}$ on the overall loss function. ADE is used to calculate the average distance between the predicted trajectories $ \hat{Y}$ and the corresponding ground truth trajectories ${Y}$. ADE plays a key role in improving the performance of distance-based metrics (more details are provided in~\cref{metrics}). 
In contrast, the primary objective of the $\mathcal{CP}$ is to reduce collisions by modeling the dynamic physical space occupied by each pedestrian, rather than relying solely on point-based representations. The $\mathcal{CP}$ consists of two main modules: Dynamic Radius Estimation for Occupied Space and Collision Penalty Calculation, which are described in the following sections.

\textbf{Dynamic Radius Estimation}
 
The Dynamic Radius Estimation for Occupied Space module aims to provide a more accurate representation of pedestrians by adapting to changes in crowd density, ranging from sparse to highly congested conditions.
In several pedestrian trajectory prediction approaches~\cite{kothari2021human, korbmacher2024toward}, individuals are commonly represented as circles with a fixed $R$ of \SI{0.2}{\meter}. This simplification is intended to enhance the realism of pedestrian representation, particularly in crowded scenarios, where physical space and interactions play a crucial role. Figure~\ref{fig:radiusexamples}a presents a visual example of empirical data on pedestrian positions, in which each circle represents the space requirement of a person. However, this fixed-radius representation does not accurately model the space requirements in high-density environments. As illustrated in~\cref{fig:radiusexamples}b and c, the radius $R$ tends to decrease as crowd density increases, often becoming smaller than \SI{0.2}{\meter}. When a constant radius of \SI{0.2}{\meter} is applied uniformly in dense crowds, it can lead to false collisions between pedestrians during model training. Figure~\ref{fig:radiusexamples}d shows that reducing the radius helps mitigate these false collisions by better aligning with the actual spacing observed on the ground. Therefore, estimating the radius $R$ of occupied area to closely match the actual physical space requirement  could enhance the proposed model's ability to predict realistic future pedestrian trajectories in various dense homogeneous and heterogeneous crowds. As a result, this module is responsible for estimating the average radius $\bar{R}$ of the occupied area of individuals within each prediction window, spanning from $t_{obs+1}$ to $t_{pred}$ time steps.

First, the set of colliding pedestrian pairs $\mathcal{O}^t$ at time step $t$ is identified by computing the Euclidean distance $d_{ij}^t$ between their 2D positions $(x_i^t, y_i^t)$ and $(x_j^t, y_j^t)$, for all $i, j \in \mathcal{N}^t$ with $i < j$. 
\begin{equation}
\mathcal{O}^t = \left\{ (i, j) \in \mathcal{N}^t \times \mathcal{N}^t \;\middle|\; i < j,\ d_{ij}^t< \SI{0.4}{\meter} \right\}.
\end{equation}
For illustration, Figure~\ref{fig:radiusexamples}c shows such overlapping pairs, highlighted by red circles and connected by red lines.


 
Afterward, the average radius of the occupied space $R^t$ in time step $t$ is computed as the mean of half the distances from center to center of all colliding pairs $\mathcal{O}^t$. The factor $1/2$ ensures that the measure is expressed as a radius per pedestrian rather than as a pairwise diameter:
\begin{equation}
R^t = \frac{1}{2|\mathcal{O}^t|} \sum_{(i, j) \in \mathcal{O}^t} d_{ij}^t,
\label{eq:avg_radius}
\end{equation}
where $|\mathcal{O}^t|$ denotes the number of pairs colliding at time $t$.

Finally, the overall average radius $\bar{R}$ over the prediction horizon is obtained by averaging $R^t$ across all time steps from $t_{\text{obs}}+1$ to $t_{\text{pred}}$:
\begin{equation}
\bar{R} = \frac{1}{t_{\text{pred}} - t_{\text{obs}}} 
\sum_{t = t_{\text{obs}}+1}^{t_{\text{pred}}} R^t.
\label{eq:final_avg_radius}
\end{equation}

In summary, averaging the radius representing the physically occupied area of pedestrians over the prediction window helps to strike a balance between collision avoidance and displacement-based error, regardless of the level of crowd density. Specifically, a smaller radius can increase the likelihood of collisions, while a larger radius can lead to higher displacement errors. Figure~\ref{fig:radiusexamples}d illustrates an example where using a radius value of \SI{0.163}{\meter} helps mitigate the false collision ratio compared to using a fixed radius of \SI{0.2}{\meter} (see~\cref{fig:radiusexamples}c).
Moreover, using a single radius value for each prediction window reduces computation time during the training process compared to assigning a separate value to each pedestrian at each time step $t$, without negatively impacting model performance.

\textbf{Collision Penalty Calculation} 

The Collision Penalty module aims to penalize predicted positions of the primary pedestrian when collisions with neighboring pedestrians occur during the prediction window. The estimated radius $\bar{R}$ plays a crucial role in the effectiveness of the $\mathcal{CP}$ module. The module operates in four steps:

\textbf{Step 1}: Calculate the pairwise distances $\mathcal{D}_i^t$ between the future position predicted $(\hat{x}_i^t, \hat{y}_i^t)$ of the pedestrian $i$ and $(\hat{x}_j^t, \hat{y}_j^t)$ of each pedestrian $j$ in time step $t$, where $j \in \mathcal{N}_i^{t}$, and $\mathcal{N}_i^{t}$ denotes the set of individuals present in the scene along with the pedestrian $i$ in time step $t$. For clarity, the Euclidean distance $d_{ij}^{t}$ between $(\hat{x}_i^t, \hat{y}_i^t)$ and each $(\hat{x}_j^t, \hat{y}_j^t)$  is calculated at the time step $t$  to construct $\mathcal{D}_i^t$. Specifically, $\mathcal{D}_i^{t}$ is defined as:

\begin{equation}
\mathcal{D}_i^{t} = \left\{ d_{ij}^{t} \,\middle|\, j \in \mathcal{N}_i^{t},\ i \neq j \right\},
\end{equation}

\textbf{Step 2:} Detect potential collisions $\mathcal{C}_i^{t}$ between the primary pedestrian $i$ and pedestrians $j \in \mathcal{N}_i^{t}$. A collision is defined based on a distance threshold $\tau$, which is set as twice the estimated radius $\bar{R}$:

\begin{equation}
\tau = 2\bar{R}.
\end{equation}

In this loss function, a collision between a pedestrian $i$ and a pedestrian $j$ is identified if the pairwise distance $d_{ij}^{t} \in \mathcal{D}_i^t$ is smaller than the threshold $\tau$. This threshold balances collision avoidance and displacement accuracy. For each detected collision, the corresponding distance $d_{ij}^{t}$ is stored as a collision weight. That is, this step selects and stores only the distances that indicate collisions (i.e., values below the threshold). 

The collision set $\mathcal{C}_i^{t}$ at time step $t$ is formally defined as:

\begin{equation}
\mathcal{C}_i^{t} = \left\{ d_{ij}^{t} \,\middle|\, d_{ij}^{t} < \tau,\ d_{ij}^{t} \in \mathcal{D}_i^t \right\}.
\end{equation}

\textbf{Step 3:} Once the collision distances $\mathcal{C}_i^{t}$ are identified, this step aims to assign penalties to each collision distance. First, the distances are normalized to a standard scale of $[0,1]$ to facilitate a consistent comparison of the severity of the collision in different scenarios. Specifically, for each collision detected between the primary pedestrian $i$ and a pedestrian $j$ at the time step $t$, the pairwise distance $d_{ij}^{t}$ is normalized by dividing it by the collision threshold $\tau$. $\widetilde{\mathcal{C}}_i^{t}$ is defined as:

\begin{equation}
\widetilde{\mathcal{C}}_i^{t} = \left\{ \frac{d_{ij}^{t}}{\tau} \,\middle|\, d_{ij}^{t} \in \mathcal{C}_i^{t} \right\}.
\end{equation}
This results in a set of normalized distances $\widetilde{\mathcal{C}}_i^{t}$ that reflect the relative severity of the collisions: values close to $0$ indicate strong (i.e. very close) collisions, while values closer to $1$ indicate less severe interactions near the threshold.
Next, the normalized distances ($\widetilde{\mathcal{C}}_i^{t}$) are converted into collision penalty values, denoted by ${\mathcal{P}}_{i}^{t}$, by applying an inverse mapping:

\begin{equation}
{\mathcal{P}}_{i}^{t} = 1 - \widetilde{\mathcal{C}}_i^{t}.
\end{equation}
 
Alternative formulation: This formulation ensures that the penalty increases with the size of the overlapping area, while distances near the threshold produce smaller penalties. 

\textbf{Step 4:} The normalized collision penalties $\mathcal{P}_{i}^{t}$ for all predicted future trajectories $\hat{Y}$ are summed to compute the total collision penalty $\mathcal{CP}$. For each predicted trajectory $\hat{Y}_i$, penalties are added in all future time steps. Then, the total penalties of all trajectories are summed to produce the final $\mathcal{CP}$, given by:

\begin{equation}
\mathcal{CP} =  \sum_{i} \sum_{t = t_{\text{obs+1}}}^{t_{\text{pred}}} \mathcal{P}_{i}^{t},
\end{equation}
  
\subsection{Evaluation Metrics}
\label{metrics}
To evaluate the performance of the proposed model and the comparison models (baseline models), two distance-based metrics~\cite{vemula2018social, zhou2024trajpred}, ADE and the final displacement error (FDE), are utilized. ADE and FDE serve as standard measures for quantifying the displacement error of the predicted trajectories. However, these metrics do not account for potential overlaps or collisions within a scene. Therefore, the Collision Rate (CR) metric~\cite{kothari2021human} is also used to measure the collision rate among predicted trajectories. These metrics are defined below.

\begin{itemize}
    \item \textbf{ADE}: Measures the average Euclidean distance between predicted and ground-truth
    trajectories over all time steps. It is defined as:

    \begin{equation}
        \text{ADE} = \frac{1}{M (t_{\text{pred}} - t_{\text{obs}})} \sum_{i=1}^{M} \sum_{t=t_{\text{obs}}+1}^{t_{\text{pred}}} \sqrt{(x_{i}^{t} - \hat{x}_{i}^{t})^2 + (y_{i}^{t} - \hat{y}_{i}^{t})^2},
    \end{equation}
where $M$ is the total number of predicted future trajectories across all scenes in the test set.

    \item \textbf{FDE}: Calculates the Euclidean distance between the predicted final position and the ground-truth final position at end of the prediction window $t_{pred}$.

    \begin{equation}
        \text{FDE} = \frac{1}{M} \sum_{i=1}^{M} \sqrt{(x_{i}^{t_{\text{pred}}} - \hat{x}_{i}^{t_{\text{pred}}})^2 + (y_{i}^{t_{\text{pred}}} - \hat{y}_{i}^{t_{\text{pred}}}
        )^2}.
\end{equation}
    
\item \textbf{CR}: Measures the percentage of predicted trajectories that result in at least one collision. 
A predicted trajectory $\hat{Y}_i$ for person $i$ is considered a collision trajectory if, at any future time step $t$, 
there is at least one collision with a neighboring person $j$ in the predicted scene. 
\Cref{eq:collision} determines whether a collision occurs between pedestrians $i$ and $j$ at a given time step $t$. For example, in \Cref{fig:CRExample}, the predicted trajectories $\hat{Y_1}$ and $\hat{Y_2}$ are considered collision trajectories 
because there is an overlap between person~$1$ and person~$2$ at time step $t_{\text{obs}+5}$. 
In contrast, $\hat{Y_3}$ does not collide with any neighbor at any time step. This metric reflects the model’s ability to learn collision avoidance behavior. 
The CR is computed as:

\begin{figure} 
\centering
\includegraphics[width=1\linewidth]{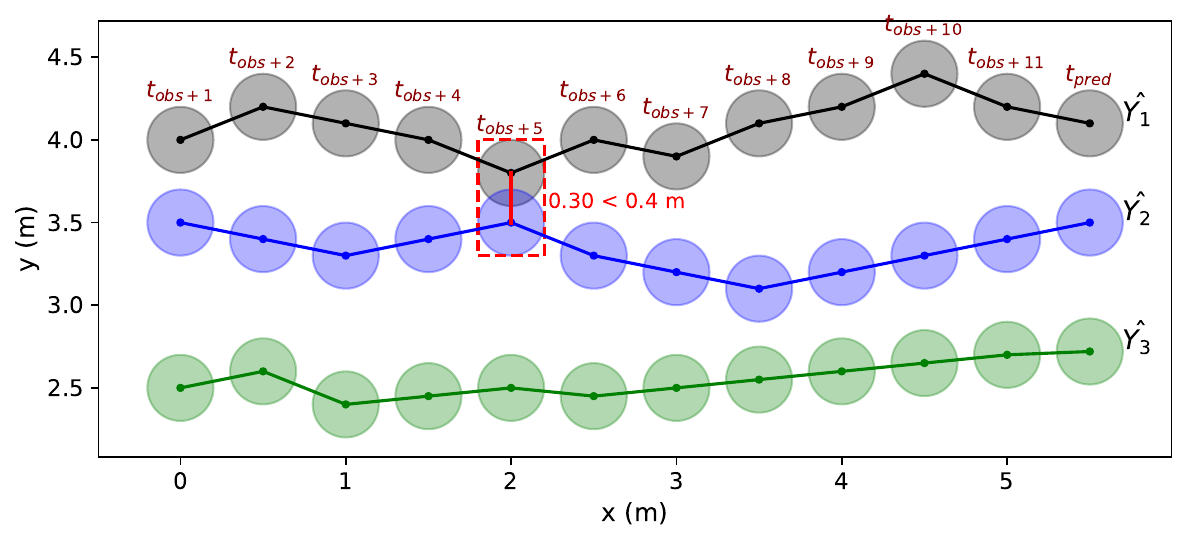}
    \caption{ Illustration of collision rate (CR) computation. Each circle represents the space occupied by a person (with radius \SI{0.2}{\meter}). 
A collision occurs when two circles overlap. In this example, $\hat{Y}_1$ and $\hat{Y}_2$ collide at $t_{\text{obs}+5}$, 
while $\hat{Y}_3$ remains collision-free. Thus, the CR for this scene is $2/3$.   
    }
    \label{fig:CRExample}
\end{figure}

\begin{equation}
\text{CR} = \frac{1}{|\mathcal{S}|} \sum_{\hat{Y} \in \mathcal{S}} \text{COL}(\hat{Y}),
\end{equation}

where $\mathcal{S}$ is the set of all scenes in the test set, and $\hat{Y} = \{\hat{Y}_1, \hat{Y}_2, \dots, \hat{Y}_n\}$ represents the predicted future trajectories of all $n$ individuals in the corresponding scene over the prediction window from $t_{\text{obs+1}}$ to $t_{\text{pred}}$. The function $\text{COL}(\hat{Y})$ is defined as:

\begin{equation}
\text{COL}(\hat{Y}) =
\frac{1}{n}
\sum_{i=1}^{n} \min\left(1,  
\sum_{\substack{j=1 \\ j \neq i}}^{n} 
\sum_{t=t_{\text{obs}}+1}^{t_{\text{pred}}}  
\left[ \sqrt{(\hat{x}_i(t) - \hat{x}_j(t))^2 + (\hat{y}_i(t) - \hat{y}_j(t))^2} < 0.4 \right] 
\right),
\label{eq:colPreTr}
\end{equation}

if we refer to $P$ as 
$\sqrt{(\hat{x}_i(t) - \hat{x}_j(t))^2 + (\hat{y}_i(t) - \hat{y}_j(t))^2} < 0.4$, 
then the Iverson bracket is defined as:

\begin{equation}
[P] = 
\begin{cases}
1, & \text{if } P \text{ is true}, \\
0, & \text{otherwise}.
\end{cases}
\end{equation}

As illustrated in~\cref{fig:CRExample}, the $\text{COL}(\hat{Y})$ in the predicted scene from $t_{\text{obs}+1}$ to $t_{\text{pred}}$ is 
$\text{COL}(\{\hat{Y}_1, \hat{Y}_2, \hat{Y}_3\}) = \frac{2}{3}$, 
since two trajectories ($\{\hat{Y}_1, \hat{Y}_2\}$) out of three result in collisions. 
Because this example contains only one scene (prediction window), the overall CR is also $\frac{2}{3}$.

\end{itemize}

\section{Experiments and Results}
\label{sec:experimentalandresults}
\subsection{Datasets}
\label{datasets}
This section aims to create the data sets required for training and evaluating the proposed model, the baseline methods, and the current approaches. Covers the data source and the necessary preparation steps to introduce the required datasets, including homogeneous datasets categorized into low density (lowD), medium density (mediumD), high density (highD) and very high density (veryHD) and a heterogeneous dataset (allD).

\subsubsection{Data Source}
\begin{figure}
\centering
\includegraphics[width=0.7\linewidth]{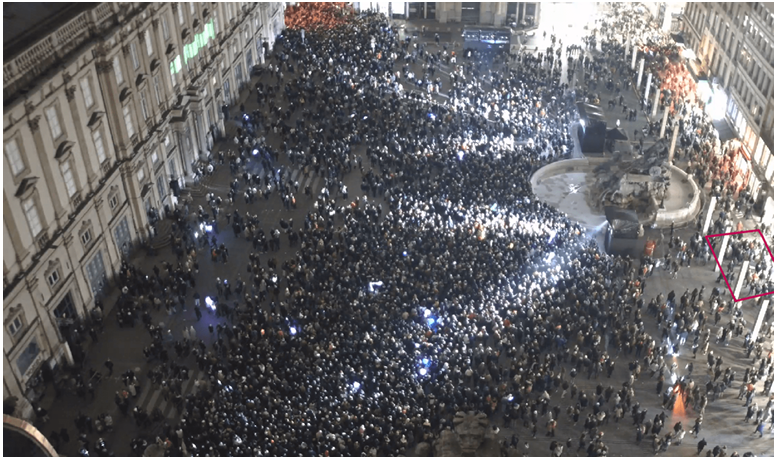}
    \caption{Top view of Lyon's Festival of Lights 2022, with the trajectory tracking region highlighted in red.  Figure from \cite{Dufour2025}.}
    \label{fig:trackingregion}
\end{figure}

The data of the pedestrian trajectory extracted and utilized in the most relevant approach (TTC-Social LSTM)~\cite{korbmacher2024toward} serves as the primary data source for the creation of the data sets. Initially, videos were recorded with a top view camera during Lyon's Festival of Lights 2022 \cite{Dufour2025}, specifically within a region $\SI{9}{\meter} \times \SI{6.5}{\meter}$ on Place des Terreaux, highlighted by the red rectangle in~\cref{fig:trackingregion}. Subsequently, the PeTrack software \cite{boltes_2025_15119517} was used to accurately extract pedestrian trajectories from recorded footage of the tracking region. The extracted pedestrian coordinates $(x,y)$ over times are measured in meters. 

A total of 5,195 individual raw trajectories were collected, with an average duration of \SI{12.38}{\second}, a mean velocity of \SI{0.62}{\meter\per\second} and an average density of \SI{0.95}{\text{pedestrians}\per\square\meter}, with three time steps per second. The trajectory data captures unidirectional and bidirectional pedestrian flows and covers a wide range of densities, from \SIrange{0.2}{2.2}{pedestrians\per\square\meter}. More details of the data can be found in Ref.~\cite{korbmacher2024toward}.

\subsubsection{Datasets Preparation}

Generating the required datasets from raw trajectory data involves two main steps: trajectory segmentation and density-based classification. These steps rely on key parameter values, including segmented trajectory length, observed trajectory length, future trajectory length, stride value, and density levels, as defined in the main baseline approach.

\begin{figure}
\centering
\includegraphics[width=1\linewidth]{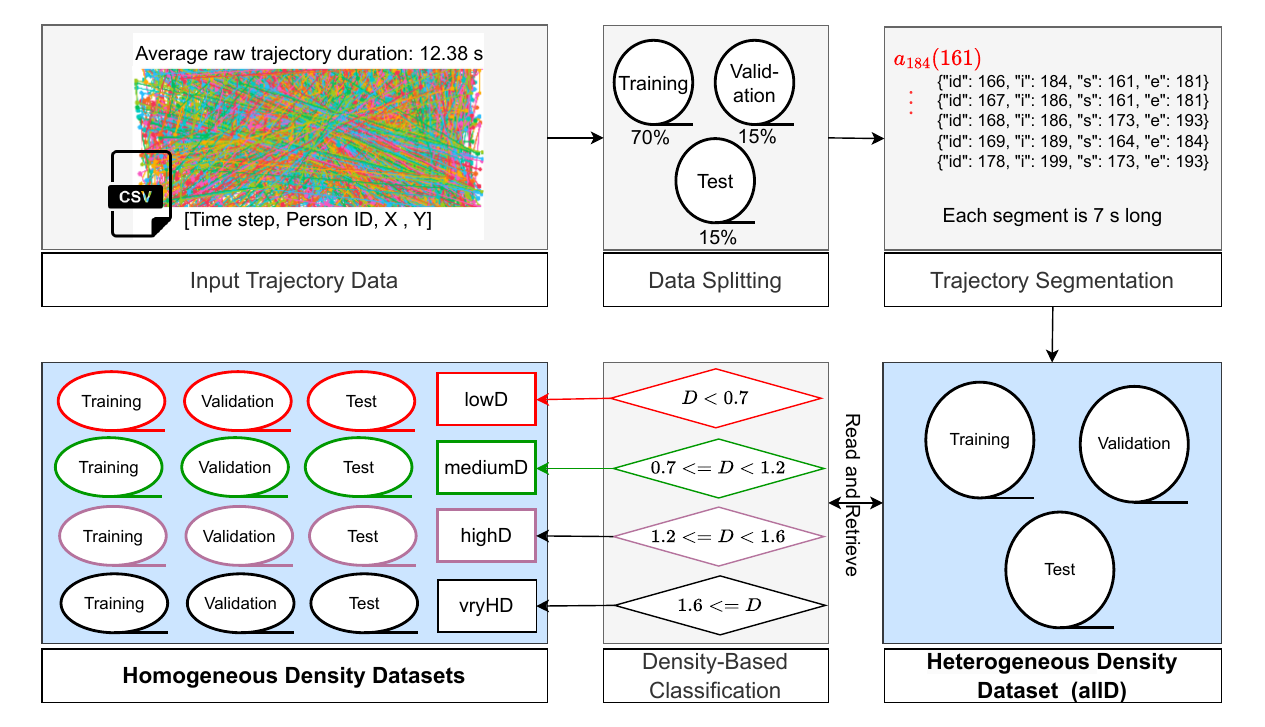}
 \caption{Dataset preparation flowchart. In the trajectory segmentation step, \( a_{184}(161) \) denotes the trajectory segment of person ID 184 starting from frame 161. Here, \( i \) refers to the person ID, \( s \) to the initial frame (start time step), and \( e \) to the final time step. Time steps are represented by frame orders. In the density classification step, \( D \) denotes the density.
}
    \label{fig:datasetpreparation}
\end{figure}

The trajectory segmentation step applies a slicing strategy with a stride $\Delta=12$ time steps to generate the segmented trajectories $\mathcal{\text{a}}_i$ from each raw trajectory  \( \mathcal{ A}_i= \{(x_i^t, y_i^t) \mid t=t_1, t_2,\dots,t_T\} \) of pedestrian \(i\), where \(T\) is the total number of time steps in the raw trajectory. Each generated segment $\mathcal{\text{a}}_i(t_1)$ of person $i$ represents $L=21$ consecutive time steps (\SI{7}{\second}) started from time step $t_1$. The slicing strategy creates overlapping between consecutive trajectory segments, making the most effective use of the raw trajectory data. The generated segments $a_i$ from the raw trajectory \(\mathcal{A}_i\) are described as:

\begin{equation}
    \text{Slicing}(\mathcal{A}_i (t_1)) = \{\mathcal{\text{a}}_i (t_{\text{init}}) \mid \text{init} = t_{1+(k-1) \times \Delta}, k=1,2, \dots K  \},
\end{equation} 
where $K$ is the total number of trajectories generated, and $K$ is determined as:

\begin{equation}
  K = \left\lfloor \frac{T - L}{\Delta} \right\rfloor + 1,  
\end{equation} 
and $a_i(t_{1})$ mathematically defined as:
\begin{equation}
a_i(t_1) = \left\{ \left( x_i^{t_{1+l}}, y_i^{t_{1+l}} \right) \mid l = 0, 1, \dots, 20 \right\}
\end{equation}

The first $9$ time steps (\SI{3}{\second}) of each segment \(a_i (t_1)\), spanning from \(t_1\) to \(t_{1+8}\), for the observed trajectory \(X_i\). The remaining $12$ time steps (\SI{4}{\second}), spanning from \(t_{1+9}\) to \(t_{1+20}\), for the future trajectory \(Y_i\). Finally, in this step, the heterogeneous density dataset (allD) is prepared as follows: 1) the raw trajectories \(\mathcal{A}_i\) are divided based on the scenes into 70\% for the training set and 15\% each for the validation and test sets, and 2) the slicing strategy is applied independently to each set to generate the trajectories \( a_i \), which represent the main samples in the sets.
 As shown in~\cref{tab:dataset_distribution}, this dataset consists of a total of $8,341$ trajectories (\( a_i \)), divided into $5,773$ for the training set, $1,317$ for the validation set, and $1,251$ for the test set. It is important to note that the splitting is performed first to ensure that there is no overlap of trajectories between the sets. 

\begin{table}[h!]
\centering
\renewcommand{\arraystretch}{1.2} 
\setlength{\tabcolsep}{6pt} 
\caption{Dataset distribution across training, validation, and test sets (trajectory). }
\begin{tabularx}{\textwidth}{>{\raggedright\arraybackslash}X>{\centering\arraybackslash}X>{\centering\arraybackslash}X>{\centering\arraybackslash}X>{\centering\arraybackslash}X}
\hline
\toprule
\textbf{Dataset} & \textbf{Training} & \textbf{Validation} & \textbf{Test} & \textbf{Total} \\ \midrule
allD & 5773 & 1317 & 1251 & 8341 \\
lowD & 440 & 101 & 96 & 637 \\ 
MediumD & 2021 & 469 & 493 & 2983 \\  
highD & 2288 & 298 & 352 & 2938 \\  
veryHD & 1024 & 449 & 310 & 1783 \\  
 \bottomrule
\end{tabularx}
\vspace{5pt} 
 
\label{tab:dataset_distribution}
\end{table}

The second step, density-based classification, aims at preparing homogeneous density datasets from allD. Classifies trajectories \(a_i\) into four density levels based on the density of the scenes that they traverse. Then it generates lowD, mediumD, highD, and veryHD data sets.
To achieve this purpose:
1) calculate the area of the scene at each time step \( t \) in \( a_i(t_1) \) using the coordinates of the trajectories of the individuals present in that scene. 
2) compute the scene density  Density($a_i^{t} $)   at each time step t within the segmented trajectory $a_i(t_1)$ by dividing the number of individuals in the scene by its area at time step t.
3) compute the average density of a trajectory \(a_i(t_1)\), which is calculated as:
\begin{equation}
\text{AverageDensity}(a_i(t_1)) = \frac{1}{21} \sum_{t=t_1}^{t_{1+20}} \text{Density}(a_i^t).
\end{equation}

4) classify the trajectory \(a_i(t_1)\) into one of the four datsets based on its average density:

\begin{equation}
\text{Class}(a_i(t_1)) =
\begin{cases} 
\text{lowD}, & \text{if } \text{AverageDensity}(a_i(t_1)) < 0.7 \, \text{ped/m}^2, \\
\text{mediumD}, & \text{if } 0.7 \leq \text{AverageDensity}(a_i(t_1)) < 1.2 \, \text{ped/m}^2, \\
\text{highD}, & \text{if } 1.2 \leq \text{AverageDensity}(a_i(t_1)) < 1.6 \, \text{ped/m}^2, \\
\text{veryHD}, & \text{if } \text{AverageDensity}(a_i(t_1)) \geq 1.6 \, \text{ped/m}^2.
\end{cases}
\end{equation}

Table~\ref{tab:dataset_distribution} presents the number of trajectories $a_i(t_1)$ in the training, validation, and test sets in the four homogeneous density data sets.

\subsection{Implementation Details}

All experiments in this study were performed on a supercomputer\footnote{JUWELS Booster is a multi-petaflop modular supercomputer operated by the Jülich Supercomputing Center at Forschungszentrum Jülich. For further details, refer to: \url{https://www.fz-juelich.de/en/ias/jsc/systems/supercomputers/juwels}} at the Jülich Supercomputing Centre using the TrajNet++ benchmark~\cite{kothari2021human} implemented in PyTorch. To ensure consistency and fair comparison, all models, including the proposed model and those used for comparison, were trained and evaluated within the same framework using identical datasets (\cref{datasets}), evaluation metrics (\cref{metrics}), and hyperparameters.

The default values were used for most hyperparameters as defined in the original implementation of the models within the benchmark framework, consistent with the baseline approach. Specifically, we used a learning rate of 0.001 with the Adam optimizer. The observation duration was set to 9 time steps and the prediction length to 12 time steps. The collision check radius was set to \SI{0.2}{\meter}. The batch size was fixed at 8. Training was carried out for up to 15 epochs or until interrupted by an early stopping mechanism with a patience of 5 epochs.

\subsection{Results and Discussion}
In this section, several experiments were conducted to evaluate the performance of the introduced model. These experiments are categorized as follows: (1) evaluation across various homogeneous density levels, (2) evaluation on a heterogeneous density distribution, (3) analysis of the impact of dynamic occupied space, and (4) comparison with state-of-the-art models.
\subsubsection{Evaluation on Homogeneous Density Scenes}

To evaluate the effectiveness of the proposed model across varying levels of homogeneous scene density, its performance was compared to that of two other models: TTC-Social LSTM and ADE-Social LSTM. The latter uses only the ADE as its loss function, without any collision penalty. It serves as the main baseline for evaluating the performance of the proposed model and TTC-Social LSTM in terms of ADE, FDE, and CR. The evaluation was carried out using four datasets, each representing a distinct and uniformly distributed density level: lowD, mediumD, highD, and veryHD.

\begin{table}[!h]
\centering
\renewcommand{\arraystretch}{1.2} 
\setlength{\tabcolsep}{5pt} 
\caption{Performance evaluation across homogeneous density levels (lowD, mediumD, highD, and veryHD). \(\textcolor{green}{\downarrow}\) indicates improvement in the corresponding metric compared to ADE-Social LSTM, while \(\textcolor{red}{\uparrow}\) indicates a decline in performance.}
\begin{tabularx}{\textwidth}{l>{\centering\arraybackslash}X>{\centering\arraybackslash}X>{\centering\arraybackslash}X>{\centering\arraybackslash}X>{\centering\arraybackslash}X>{\centering\arraybackslash}X>{\centering\arraybackslash}X>{\centering\arraybackslash}X}
\toprule
\multirow{2}{*}{\textbf{Model}} & \multicolumn{2}{c}{\textbf{lowD}} & \multicolumn{2}{c}{\textbf{mediumD}} & \multicolumn{2}{c}{\textbf{highD}} & \multicolumn{2}{c}{\textbf{veryHD}} \\ 
 & \textbf{Value} & \textbf{Difference} & \textbf{Value} & \textbf{Difference} & \textbf{Value} & \textbf{Difference} & \textbf{Value} & \textbf{Difference} \\ \midrule 
\textbf{ADE-Social LSTM } & \multicolumn{8}{c}{} \\  
   
ADE (m) & 0.499 & -- & 0.345 & -- & 0.241 & -- & 0.259 & -- \\  
FDE (m) & 0.949 & -- & 0.671 & -- & 0.418 & -- & 0.456 & -- \\  
CR (\%) & 40.62 & -- & 29.21 & -- & 33.8 & -- & 51.29 & -- \\ \midrule
\textbf{TTC-Social LSTM} & \multicolumn{8}{c}{} \\  
Optimal $\lambda$ & 0.001 & -- & 0.002 & -- & 0.001 & -- & 0.002 & -- \\ 
ADE (m) & 0.469 & \(-0.03\ \textcolor{green}{\downarrow}\) & 0.307 & \(-0.038\ \textcolor{green}{\downarrow}\) & 0.251 & \(0.01\ \textcolor{red}{\uparrow}\) & 0.319 & \(0.06\ \textcolor{red}{\uparrow}\) \\  
FDE (m) & 0.904 & \(-0.045\ \textcolor{green}{\downarrow}\) & 0.549 & \(-0.122\ \textcolor{green}{\downarrow}\) & 0.435 & \(0.017\ \textcolor{red}{\uparrow}\) & 0.577 & \(0.121\ \textcolor{red}{\uparrow}\) \\  
CR (\%) & 37.5 & \(-3.12\ \textcolor{green}{\downarrow}\) & 20.6 & \(-8.61\ \textcolor{green}{\downarrow}\) & 19.3 & \(-14.5\ \textcolor{green}{\downarrow}\) & 38.7 & \(-12.59\ \textcolor{green}{\downarrow}\) \\ \midrule 
\textbf{Our model} & \multicolumn{8}{c}{} \\  
Optimal $\lambda$ & 0.01 & -- & 0.01 & -- & 0.001 & -- & 0.002 & -- \\  
ADE (m) & 0.463 & \(-0.036\ \textcolor{green}{\downarrow}\) & 0.323 & \(-0.022\ \textcolor{green}{\downarrow}\) & 0.239 & \(-0.002\ \textcolor{green}{\downarrow}\) & 0.238 & \(-0.021\ \textcolor{green}{\downarrow}\) \\ 
FDE (m) & 0.876 & \(-0.073\ \textcolor{green}{\downarrow}\) & 0.621 & \(-0.05\ \textcolor{green}{\downarrow}\) & 0.413 & \(-0.005\ \textcolor{green}{\downarrow}\) & 0.42 & \(-0.036\ \textcolor{green}{\downarrow}\) \\  
CR (\%) & 22.9 & \(-17.72\ \textcolor{green}{\downarrow}\) & 12.3 & \(-16.91\ \textcolor{green}{\downarrow}\) & 25.5 & \(-8.3\ \textcolor{green}{\downarrow}\) & 47.4 & \(-3.89\ \textcolor{green}{\downarrow}\) \\ \bottomrule
\end{tabularx}

\label{tab:performance_comparison}
\end{table}

\begin{figure}[!h] 
\centering
\includegraphics[width=0.9\linewidth]{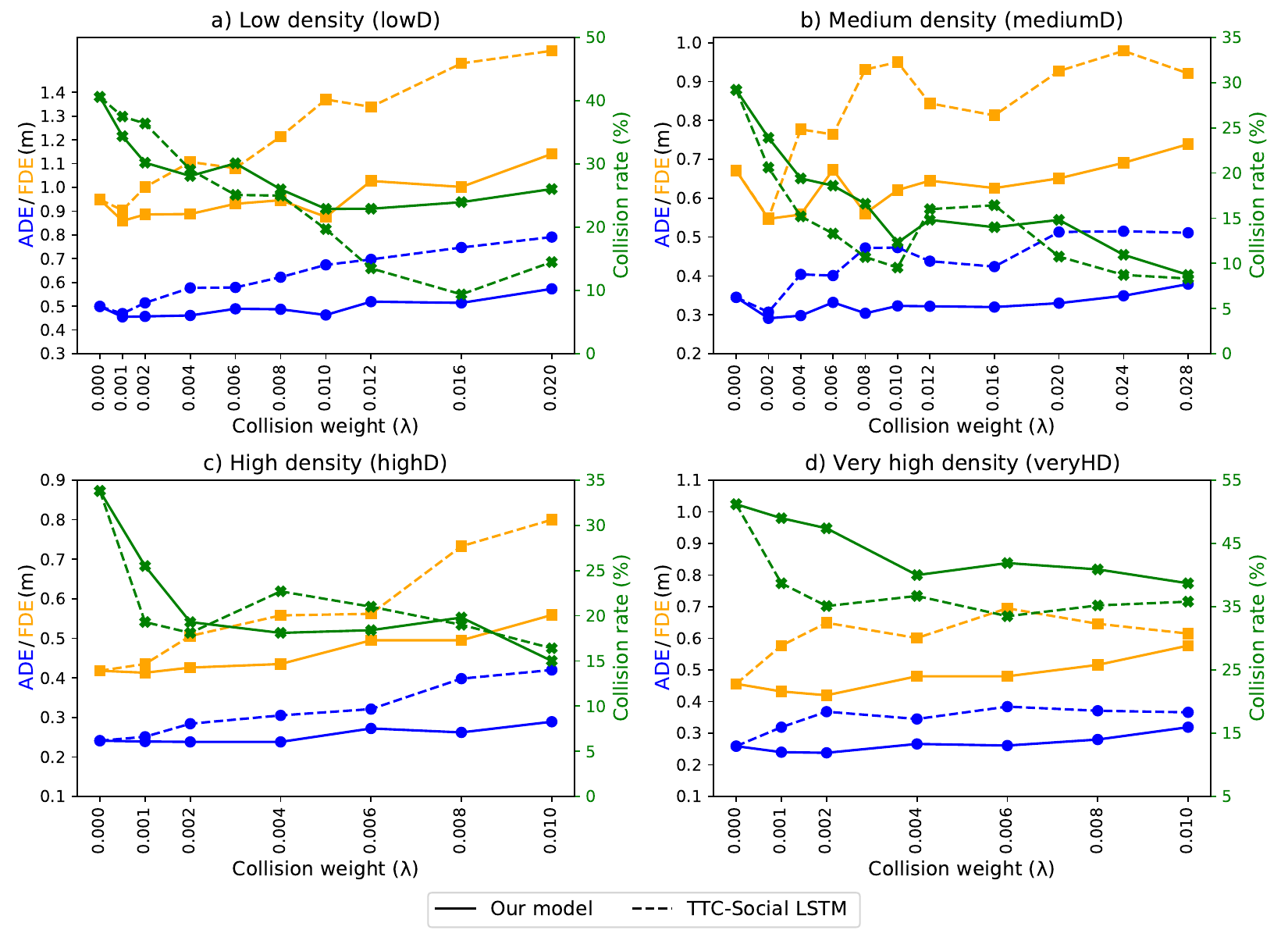}
    \caption{ Impact of collision weight ($\lambda$) on models' performance in terms of ADE, FDE, and CR across four scene density levels: a) lowD, b) mediumD, c) highD, and d) veryHD. The case $\lambda = 0$ corresponds to ADE-Social LSTM, where the collision penalty is not applied.
}
    \label{fig:lambdaOnHOmo}
\end{figure}

As shown in~\cref{tab:performance_comparison}, the new model consistently outperformed the ADE-Social LSTM across all density levels in terms of ADE, FDE, and CR. For instance, in the lowD dataset, it achieved a significant CR reduction of \SI{17.72}{\percent}, along with improvements of \SI{3.6}{\centi\meter} in ADE and \SI{7.3}{\centi\meter} in FDE. Consistently, improvements in ADE, FDE, and CR were observed in the mediumD, highD, and veryHD scenarios.

The comparison between TTC-Social LSTM and our model demonstrated the superiority of the proposed model, particularly in reducing CR, while enhancing displacement accuracy across all density levels. In contrast, TTC-Social LSTM struggled to reduce CR without increasing ADE and FDE, especially in the highD and veryHD datasets.
Furthermore, in the lowD setting, the introduced model achieved a substantial reduction \SI{17.72}{\percent} in CR, compared to \SI{3.12}{\percent} with TTC-Social LSTM, while also achieving better ADE and FDE values. In the mediumD dataset, although TTC-Social LSTM slightly outperformed the proposed model in displacement errors, the developed model obtained a much greater reduction in CR, \SI{16.91}{\percent} versus \SI{8.61}{\percent}.

The collision weight $\lambda$ used in the loss functions of our model and the TTC-Social LSTM was experimentally tuned to minimize CR without increasing either ADE or FDE (see~\cref{eq:DOC-SLSTM}).
Table~\ref{tab:performance_comparison} presents the optimal values $\lambda$ for each data set for both the presented model and TTC-Social LSTM. A higher $\lambda$ typically leads to a lower CR but often at the cost of increased displacement errors. Therefore, it is crucial to adapt $\lambda$ to maintain a balance between reducing CR and preserving the performance of ADE and FDE.
Figure~\ref{fig:lambdaOnHOmo} illustrates the impact of different values $\lambda$ on ADE, FDE, and CR for both models. In particular, the case where $\lambda = 0$ corresponds to ADE-Social LSTM, indicating the absence of a collision penalty in the loss function.

In summary, the new model successfully reduces collisions and improves displacement accuracy across all homogeneous density levels.

\subsubsection{Evaluation on Heterogeneous Density Scenes
}

To assess the performance of the proposed model under diverse pedestrian density conditions, it was trained and evaluated alongside ADE-Social LSTM and TTC-Social LSTM using the heterogeneous dataset (allD).

\begin{table}[!h]
\centering
\renewcommand{\arraystretch}{1.2} 
\setlength{\tabcolsep}{4pt} 
\caption{
Performance evaluation across heterogeneous density levels (allD). \(\textcolor{green}{\downarrow}\) indicates improvement in a metric, while \(\textcolor{red}{\uparrow}\) indicates a decline.  ``diff.'' shows the difference between ADE-Social LSTM and the other two models: the new model and TTC-Social LSTM.
}
\begin{tabular}{lcccccccccc}
\toprule
\multirow{2}{*}{\textbf{Model}} & \multicolumn{2}{c}{\textbf{allD}} & \multicolumn{2}{c}{\textbf{lowD}} & \multicolumn{2}{c}{\textbf{mediumD}} & \multicolumn{2}{c}{\textbf{highD}} & \multicolumn{2}{c}{\textbf{veryHD}} \\ 
 & \textbf{Value} & \textbf{Diff.$^*$} & \textbf{Value} & \textbf{Diff.} & \textbf{Value} & \textbf{Diff.} & \textbf{Value} & \textbf{Diff.} & \textbf{Value} & \textbf{Diff.} \\ \midrule
\makecell[l]{\textbf{ADE-Social LSTM} } & \multicolumn{10}{c}{} \\ 
ADE (m) & 0.257 & -- & 0.402 & -- & 0.281 & -- & 0.224 & -- & 0.212 & -- \\ 
FDE (m) & 0.473 & -- & 0.769 & -- & 0.538 & -- & 0.396 & -- & 0.367 & -- \\ 
CR (\%) & 39.3 & -- & 41.6 & -- & 29 & -- & 36.9 & -- & 57.4 & -- \\ \midrule
\makecell[l]{\textbf{TTC-Social LSTM} \\ (Optimal $\lambda$: 0.001)} & \multicolumn{10}{c}{} \\ 
ADE (m) & 0.288 & \(+0.031\ \textcolor{red}{\uparrow}\) & 0.445 & \(+0.043\ \textcolor{red}{\uparrow}\) & 0.321 & \(+0.04\ \textcolor{red}{\uparrow}\) & 0.258 & \(+0.034\ \textcolor{red}{\uparrow}\) & 0.221 & \(+0.009\ \textcolor{red}{\uparrow}\) \\ 
FDE (m) & 0.532 & \(+0.06\ \textcolor{red}{\uparrow}\) & 0.85 & \(+0.081\ \textcolor{red}{\uparrow}\) & 0.612 & \(+0.074\ \textcolor{red}{\uparrow}\) & 0.464 & \(+0.068\ \textcolor{red}{\uparrow}\) & 0.386& \(+0.019\ \textcolor{red}{\uparrow}\) \\ 
CR (\%) & 26.1 & \(-13.2\ \textcolor{green}{\downarrow}\) & 31.3 & \(-10.3\ \textcolor{green}{\downarrow}\) & 16.6 & \(-12.4\ \textcolor{green}{\downarrow}\) & 21.6 & \(-15.3\ \textcolor{green}{\downarrow}\) & 44.8 & \(-12.6\ \textcolor{green}{\downarrow}\) \\ \midrule
\makecell[l]{\textbf{Our model} \\(Optimal $\lambda$: 0.003)} & \multicolumn{10}{c}{} \\ 
ADE (m) & 0.248 & \(-0.01\ \textcolor{green}{\downarrow}\) & 0.368 & \(-0.034\ \textcolor{green}{\downarrow}\) & 0.274 & \(-0.007\ \textcolor{green}{\downarrow}\) & 0.219 & \(-0.005\ \textcolor{green}{\downarrow}\) & 0.203 & \(-0.009\ \textcolor{green}{\downarrow}\) \\ 
FDE (m) & 0.445 & \(-0.028\ \textcolor{green}{\downarrow}\) & 0.678 & \(-0.091\ \textcolor{green}{\downarrow}\) & 0.507 & \(-0.031\ \textcolor{green}{\downarrow}\) & 0.376 & \(-0.02\ \textcolor{green}{\downarrow}\) & 0.355 & \(-0.012\ \textcolor{green}{\downarrow}\) \\ 
CR (\%) & 29.9 & \(-9.4\ \textcolor{green}{\downarrow}\) & 35.4 & \(-6.2\ \textcolor{green}{\downarrow}\) & 21.7 & \(-7.3\ \textcolor{green}{\downarrow}\) & 26.1 & \(-10.8\ \textcolor{green}{\downarrow}\) & 45.8 & \(-11.6\ \textcolor{green}{\downarrow}\) \\ \hline
\end{tabular}

\label{tab:performance_comparison_allD}
\end{table}

\begin{figure}[!h]
\centering
\includegraphics[width=0.6\linewidth]{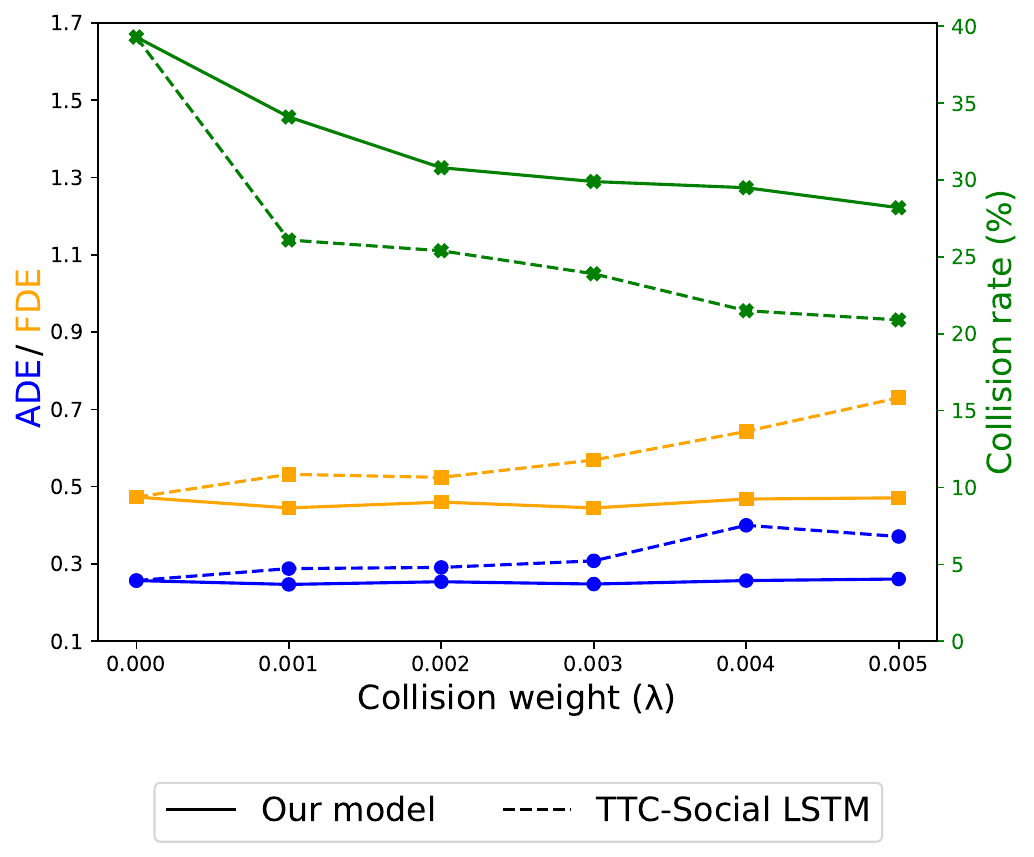}
    \caption{
    Impact of varying the collision weight ($\lambda$) on ADE (m), FDE (m), and CR (\%) using the heterogeneous dataset (allD). When ($\lambda$) = 0, the setup corresponds to ADE-Social LSTM.
}
    \label{fig:mixeddensitylevels}
\end{figure}

As shown in~\cref{tab:performance_comparison_allD}, the proposed model consistently improved CR, ADE and FDE in mixed density settings. In contrast, while TTC-Social LSTM reduced the CR, it failed to maintain the displacement accuracy, as both ADE and FDE increased. In particular, the proposed model decreased the CR by \SI{9.4}{\percent} while simultaneously reducing displacement errors. On the other hand, TTC-Social LSTM improved CR by \SI{13.2}{\percent}, but this came at the cost of increasing displacement errors.
Moreover, the introduced model trained on the allD dataset demonstrated strong generalization when evaluated on the individual homogeneous density test sets: lowD, mediumD, highD, and veryHD. It consistently reduced CR and displacement errors across all test scenarios. In contrast, the TTC-Social LSTM model failed to achieve reductions in displacement errors for any of the test sets. The optimal collision weights were experimentally determined to be \num{0.01} for our model and \num{0.001} for TTC-Social LSTM (see~\cref{fig:mixeddensitylevels}).

\begin{figure}
\centering
\includegraphics[width=1\linewidth]{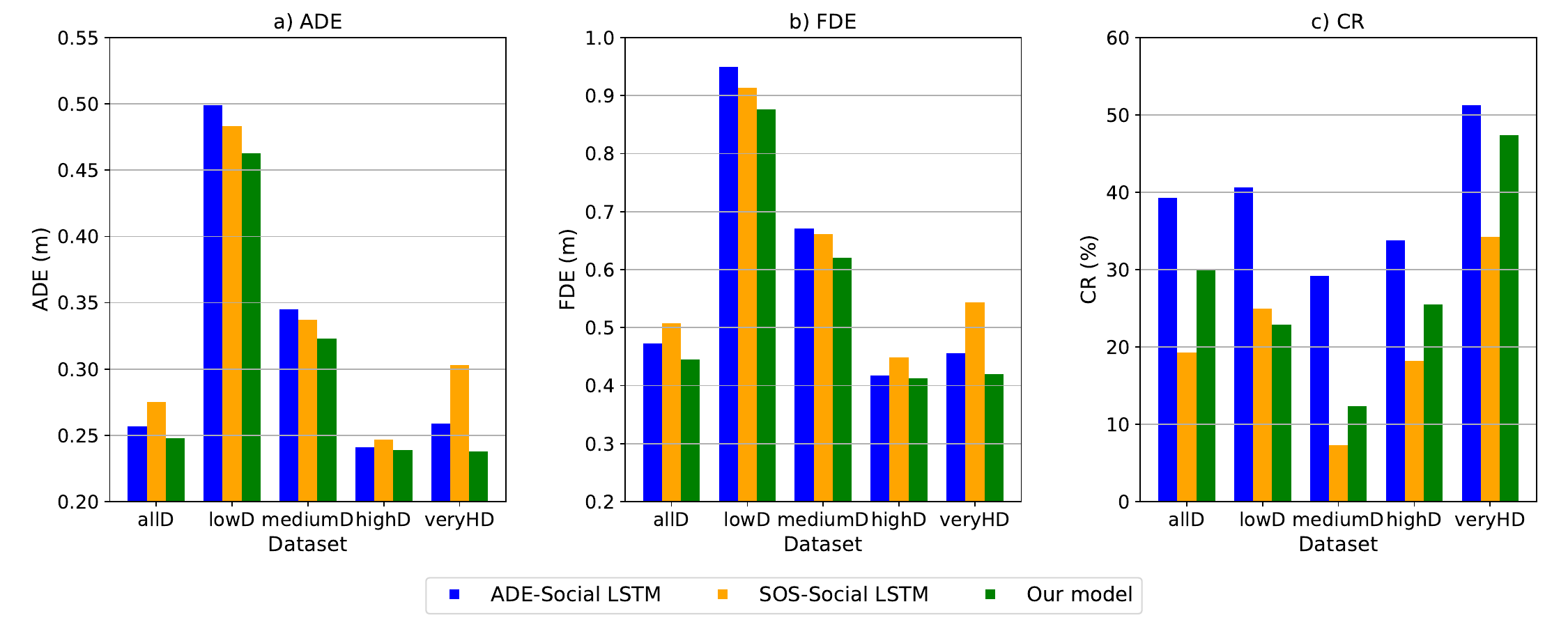}
    \caption{Impact of dynamic occupied space mechanism on the performance of the proposed model across heterogeneous (allD) and homogeneous (lowD, mediumD, highD, veryHD) crowd density scenarios.}
    \label{fig:DOShomogeneous}
\end{figure}

These results highlight the effectiveness of the model in simultaneously reducing the CR and improving displacement accuracy under heterogeneous density conditions. This performance is primarily attributed to the new loss function. As a result, the model emerges as a strong candidate for predicting real-world pedestrian trajectory in complex and dynamic crowd environments.

\subsubsection{Impact of Dynamic Occupied Space on Social LSTM Performance
}

A key contribution of the proposed loss function is its ability to dynamically estimate the radius of the space occupied by pedestrians based on the scene’s density. 
This section aims to investigate how the incorporation of this dynamically occupied space affects the performance of the proposed model.

To achieve this purpose, we introduced a new baseline model called Static Occupied Space Social LSTM (SOS-Social LSTM), which differs from the proposed model only in its use of a fixed occupied space with a radius of \SI{0.2}{\meter}, regardless of scene density. This static representation of occupied space has been used in several models in the literature, such as those in Refs.~\cite{kothari2021human, korbmacher2024toward}. The baseline model was trained and evaluated using the same datasets, parameters, and experimental setup as the proposed model.
Figure~\ref{fig:DOShomogeneous} compares the performance of ADE-Social LSTM, SOS-Social LSTM, and our model across heterogeneous (allD) and homogeneous (lowD, mediumD, highD, veryHD) datasets. The results clearly demonstrated the effectiveness of dynamic spatial modeling in improving CR, ADE and FDE across all density conditions.
In contrast, the SOS-Social LSTM failed to achieve this improvement in the allD, highD, and veryHD datasets. For example, SOS-Social LSTM reduces the CR in these three datasets by  \SI{23.04}{\percent},  \SI{15.63}{\percent}, and \SI{14.1}{\percent}, respectively (see Figure~\ref{fig:DOShomogeneous}c). However, this reduction is accompanied by an increase in displacement errors, as shown in Figures~\ref{fig:DOShomogeneous}a,b.

Additionally, under the lowD and mediumD datasets, SOS-Social LSTM reduced the CR while also improving displacement accuracy. This suggests that a fixed occupied space radius of \SI{0.2}{\meter} could be suitable for low- and medium-density environments. Despite these results, the introduced outperformed SOS-Social LSTM in the lowD setting by achieving a \SI{2.8}{\percent} lower collision rate, a \SI{2}{\centi\meter} lower ADE, and approximately \SI{4}{\centi\meter} lower FDE. In the mediumD dataset, the new model achieved better ADE and FDE, while SOS-Social LSTM outperformed the proposed model in terms of collision rate. 
The main reason for this result is that, in high-density and heterogeneous scenes, a fixed radius becomes inefficient because the actual space occupied by pedestrians is often smaller than \SI{0.2}{\meter} in crowded conditions. Therefore, adapting the radius to reflect the realistic spacing of the crowd helps balance collision avoidance with displacement accuracy.

In summary, incorporating a dynamic occupied space into the loss function enables the proposed model to adapt to varying crowd densities. This adaptive approach successfully reduces collisions and improves displacement accuracy across both homogeneous settings—from low to high density—and heterogeneous environments. In contrast, the static occupied space fails to minimize collisions and displacement errors simultaneously, especially in dense and heterogeneous scenes.

\subsection{Comparison with State-of-the-Art Approaches 
}

For further evaluation, the proposed model is compared with two state-of-the-art methods on the homogeneous and heterogeneous datasets. These approaches are Vanilla LSTM~\cite{hochreiter1997long} and Social-GAN~\cite{gupta2018social}.

The results illustrated in~\cref{tab:trajnet_comparison} demonstrated that the new model significantly reduced the CR across all datasets compared to Vanilla LSTM and Social-GAN. On the lowD dataset, it lowered CR by \SI{22.9}{\percent} and \SI{15.6}{\percent} relative to Vanilla LSTM and Social GAN, respectively. Even in the challenging veryHD setting, where managing dense crowds is difficult, the proposed model achieved a CR of only \SI{47.4}{\percent}, clearly outperforming Vanilla LSTM (\SI{61.6}{\percent}) and Social GAN (\SI{53.9}{\percent}). In addition to its advantage in CR, our model performed as well as—or better than—the other models in terms of displacement accuracy (ADE and FDE) on most homogeneous datasets. It achieved lower displacement errors than both approaches in the lowD and veryHD datasets, and performed similarly to Vanilla LSTM in veryHD while still outperforming Social GAN. On the mediumD dataset, the introduced showed slightly higher displacement errors compared to Vanilla LSTM and Social GAN. However, it still achieved a major improvement in CR, reducing it by more than \SI{20}{\percent} compared to both approaches. This highlighted the model’s strength in balancing collision reduction with prediction accuracy.
Moreover, on the allD (heterogeneous) dataset, the proposed model achieved a strong balance between displacement accuracy and collision rate. It reduced CR by \SI{14.7}{\percent} and \SI{9.6}{\percent}, while also improving displacement accuracy compared to Vanilla LSTM and Social GAN, respectively.

To sum up, our model significantly outperformed Vanilla LSTM and Social GAN in reducing collision instances across all homogeneous and heterogeneous datasets, while also achieving better displacement accuracy in most settings.
This improvement is mainly attributed to the model's ability to explicitly learn the influence of nearby pedestrians and realistic collision avoidance.

\begin{table}[!ht]
    \centering
\footnotesize	    \caption{Comparison with state-of-the-art approaches. \textit{Diff.} refers to the difference between the introduced model and the corresponding method. A \textcolor{green}{\(\downarrow\)} indicates that the compared method performs better in the given metric, while a \textcolor{red}{\(\uparrow\)} indicates that our model LSTM performs better.}
    \renewcommand{\arraystretch}{1.2}
    \setlength{\tabcolsep}{5pt}
    \begin{tabularx}{\textwidth}{l>{\centering\arraybackslash}X>{\centering\arraybackslash}X>{\centering\arraybackslash}X>{\centering\arraybackslash}X>{\centering\arraybackslash}X>{\centering\arraybackslash}X}  
        \toprule 
        
        \textbf{Approach} & \textbf{Metric} & 
\makecell{\textbf{lowD} \\ Value (Diff.$^*$)} & 
\makecell{\textbf{mediumD} \\ Value (Diff.$^*$)} & 
\makecell{\textbf{highD} \\ Value (Diff.$^*$)} & 
\makecell{\textbf{VeryHD} \\ Value (Diff.$^*$)} & 
\makecell{\textbf{allD} \\ Value (Diff.$^*$)} \\
\midrule
        \multirow{3}{*}{\textbf{Our model}} & ADE (m) & 0.46 & 0.32 & 0.24 & 0.24 & 0.25 \\
                                         & FDE (m) & 0.88 & 0.62 & 0.41 & 0.42 & 0.45 \\
                                         & CR (\%) & 22.9 & 12.3 & 25.5 & 47.4 & 29.9 \\
        \midrule
        \multirow{3}{*}{\textbf{Vanilla LSTM}} & ADE (m) & 0.47 (0.01 \textcolor{red}{$\uparrow$}) & 0.30 (-0.02 \textcolor{green}{$\downarrow$}) & 0.25 (0.01 \textcolor{red}{$\uparrow$}) & 0.24 (0.00 =) & 0.28 (0.03 \textcolor{red}{$\uparrow$}) \\
                                  & FDE (m) & 0.92 (0.04 \textcolor{red}{$\uparrow$}) & 0.56 (-0.06 \textcolor{green}{$\downarrow$}) & 0.45 (0.04 \textcolor{red}{$\uparrow$}) & 0.42 (0.00 =) & 0.53 (0.08 \textcolor{red}{$\uparrow$}) \\
                                  & CR (\%) & 45.8 (22.9 \textcolor{red}{$\uparrow$}) & 33.2 (20.9 \textcolor{red}{$\uparrow$}) & 42.0 (16.5 \textcolor{red}{$\uparrow$}) & 61.6 (14.2 \textcolor{red}{$\uparrow$}) & 44.6 (14.7 \textcolor{red}{$\uparrow$}) \\
        \midrule
        \multirow{3}{*}{\textbf{Social GAN}} & ADE (m) & 0.46 (0.00 = ) & 0.29 (-0.03 \textcolor{green}{$\downarrow$}) & 0.26 (0.02 \textcolor{red}{$\uparrow$}) & 0.28 (0.04 \textcolor{red}{$\uparrow$}) & 0.28 (0.03 \textcolor{red}{$\uparrow$}) \\
                              & FDE (m) & 0.90 (0.02 \textcolor{red}{$\uparrow$}) & 0.55 (-0.07 \textcolor{green}{$\downarrow$}) & 0.47 (0.06 \textcolor{red}{$\uparrow$}) & 0.49 (0.07 \textcolor{red}{$\uparrow$}) & 0.51 (0.07 \textcolor{red}{$\uparrow$}) \\
                              & CR (\%) & 38.5 (15.6 \textcolor{red}{$\uparrow$}) & 32.8 (20.5 \textcolor{red}{$\uparrow$}) & 34.4 (8.9 \textcolor{red}{$\uparrow$}) & 53.9 (6.5 \textcolor{red}{$\uparrow$}) & 39.5 (9.6 \textcolor{red}{$\uparrow$}) \\
        \bottomrule
    \end{tabularx}
    \label{tab:trajnet_comparison}
\end{table}

\section{Conclusion and Future Works}
\label{sec:conclusion} 
This paper introduced  a novel deep learning model designed to predict accurate, realistic, and collision-free
pedestrian trajectories in dynamic and crowded environments. By integrating a new DOS-based loss
function into the Social LSTM architecture, the proposed model explicitly learns to avoid pedestrian collisions while improving
high displacement accuracy. The proposed loss function adaptively adjusts the occupied space radius by pedestrians based on the scene
density, allowing the model to generalize across a broad range of crowd conditions, from sparse to dense crowds and from homogeneous to heterogeneous density distributions.
The proposed model was trained and evaluated on five datasets derived from real pedestrian trajectories, including four homogeneous density levels (low, medium, high, and very high density) and one heterogeneous scenario. The experimental findings showed that the model successfully reduced the number of collision instances while improving the accuracy of the displacement across all datasets. Furthermore, it consistently outperformed five baselines and state-of-the-art models in terms of both collision rate and displacement errors in most experiments.

While the proposed model effectively captures human–human interactions, it does not explicitly account for the influence of environmental constraints on pedestrian movement. To enhance its generalizability across varied environments, future work will focus on extending the model to learn human-environment interactions. This will enable the model to reflect real-world behavior in complex and structured spaces with more accuracy.

\textbf{Data and Code Availability}

The raw trajectory data used in this study is available at the following website: \url{https://madras-data-app.streamlit.app/} (accessed on 1 April 2024). In addition, the TrajNet++ benchmark, which was used in the implementation of this model and for dataset construction, is described at: \url{https://thedebugger811.github.io/posts/2020/03/intro_trajnetpp/}.
The code implementing the DOS-based loss function is available upon request.
 
\textbf{Acknowledgments} \\
The authors sincerely thank Antoine Tordeux and Raphael Korbmacher for their insightful discussions and for providing us with the trajectory datasets and the TTC-Loss function code. 

\textbf{Conflicts of Interest} \\
The authors declare no conflict of interest.

\bibliographystyle{unsrt}
\bibliography{main.bib}

\end{document}